\documentclass[letterpaper, 10 pt, journal, twoside]{IEEEtran}
\usepackage{cite}
\usepackage{amsmath,amssymb,amsfonts}
\usepackage{algorithmic}
\usepackage{graphicx}
\usepackage{textcomp}
\usepackage{xcolor}
\usepackage{multirow}
\usepackage{threeparttable}
\usepackage{array}
\usepackage{booktabs}
\usepackage{multirow}
\usepackage{amssymb}
\usepackage{amsfonts}
\usepackage{amsbsy}
\ifCLASSINFOpdf
\else
\fi
\begin{document}
%
\title{Semi-Elastic LiDAR-Inertial Odometry}
%
%
%

\author{Zikang~Yuan$^{1}$, Fengtian~Lang$^{2}$, Tianle~Xu$^{2}$, Ruiye~Ming$^{2}$, Chengwei~Zhao$^{3}$ and Xin~Yang$^{2}$
	\thanks{This work was not supported by any organization.}
	\thanks{$^{1}$Zikang~Yuan is with Institute of Artificial Intelligence, Huazhong University of Science and Technology, Wuhan, 430074, China. (E-mail: {\tt\small yzk2020@hust.edu.cn})}%
	\thanks{$^{2}$Fengtian~Lang, Tianle~Xu, Ruiye~Ming and Xin~Yang are with the Electronic Information and Communications, Huazhong University of Science and Technology, Wuhan, 430074, China. (E-mail: {\tt\small M202372913@hust.edu.cn; tianlexu@hust.edu.cn; M202272555@hust.edu.cn; xinyang2014@hust.edu.cn})}%
	\thanks{$^{3}$Chengwei Zhao is with Hangzhou Guochen Robot Technology Company Limited, Hangzhou, 311200, China. (E-mail: {\tt\small chengweizhao0427@gmail.com})}%
}
%
%

\markboth{IEEE Robotics and Automation Letters. Preprint Version. Accepted Month, Year}
{FirstAuthorSurname \MakeLowercase{\textit{et al.}}: ShortTitle} 

%



\maketitle

\begin{abstract}
Existing LiDAR-inertial state estimation assumes that the state at the beginning of current sweep is identical to the state at the end of last sweep. However, if the state at the end of last sweep is not accurate, the current state cannot satisfy the constraints from LiDAR and IMU consistently, ultimately resulting in local inconsistency of solved state (e.g., zigzag trajectory or high-frequency oscillating velocity). This paper proposes a semi-elastic optimization-based LiDAR-inertial state estimation method, which imparts sufficient elasticity to the state to allow it be optimized to the correct value. This approach can preferably ensure the accuracy, consistency, and robustness of state estimation. We incorporate the proposed LiDAR-inertial state estimation method into an optimization-based LiDAR-inertial odometry (LIO) framework. Experimental results on four public datasets demonstrate that: 1) our method outperforms existing state-of-the-art LiDAR-inertial odometry systems in terms of accuracy; 2) semi-elastic optimization-based LiDAR-inertial state estimation can better ensure consistency and robustness than traditional and elastic optimization-based LiDAR-inertial state estimation. We have released the source code of this work for the development of the community.
\end{abstract}


%
\IEEEpeerreviewmaketitle

\section{Introduction}
\label{Introduction}

\IEEEPARstart{3}{D} light detection and ranging (LiDAR) has become the commonly used sensor in the fields of robotics and autonomous driving. In theory, LiDAR-only odometry \cite{zhang2017low, wang2021f, wang2020intensity, deschaud2018imls, dellenbach2022ct} can achieve real-time pose estimation and transform points collected at different times into a unified coordinate system to obtain a global map. However, the Iterative Closest Point (ICP) algorithm used in LiDAR odometry can hardly solve a pose fastly without a reliable initial motion value. Meanwhile, scenes without enough geometric features fail to provide reliable constraints for pose estimation. Introducing Inertial Measurement Unit (IMU) can effectively address these issues with minimal memory and time consumption.

Existing ICP-based pose estimation methods in LiDAR odometry can be mainly divided into two categories: 1) the traditional ICP algorithm, which is widely used and exemplified by LOAM \cite{zhang2017low}. This method optimizes the pose only at the end of a sweep; 2) the elastic ICP algorithm exemplified by CT-ICP \cite{dellenbach2022ct}, which enables the state at the beginning of current sweep to be adjusted, and in turn reduces the negative impact of the last state estimation result on current state estimation. However, both two ICP algorithms encounter challenges after being integrated with IMU pre-integration to the optimization-based LiDAR-inertial odometry (LIO) framework.

\begin{figure}
	\begin{center}
		\includegraphics[scale=0.435]{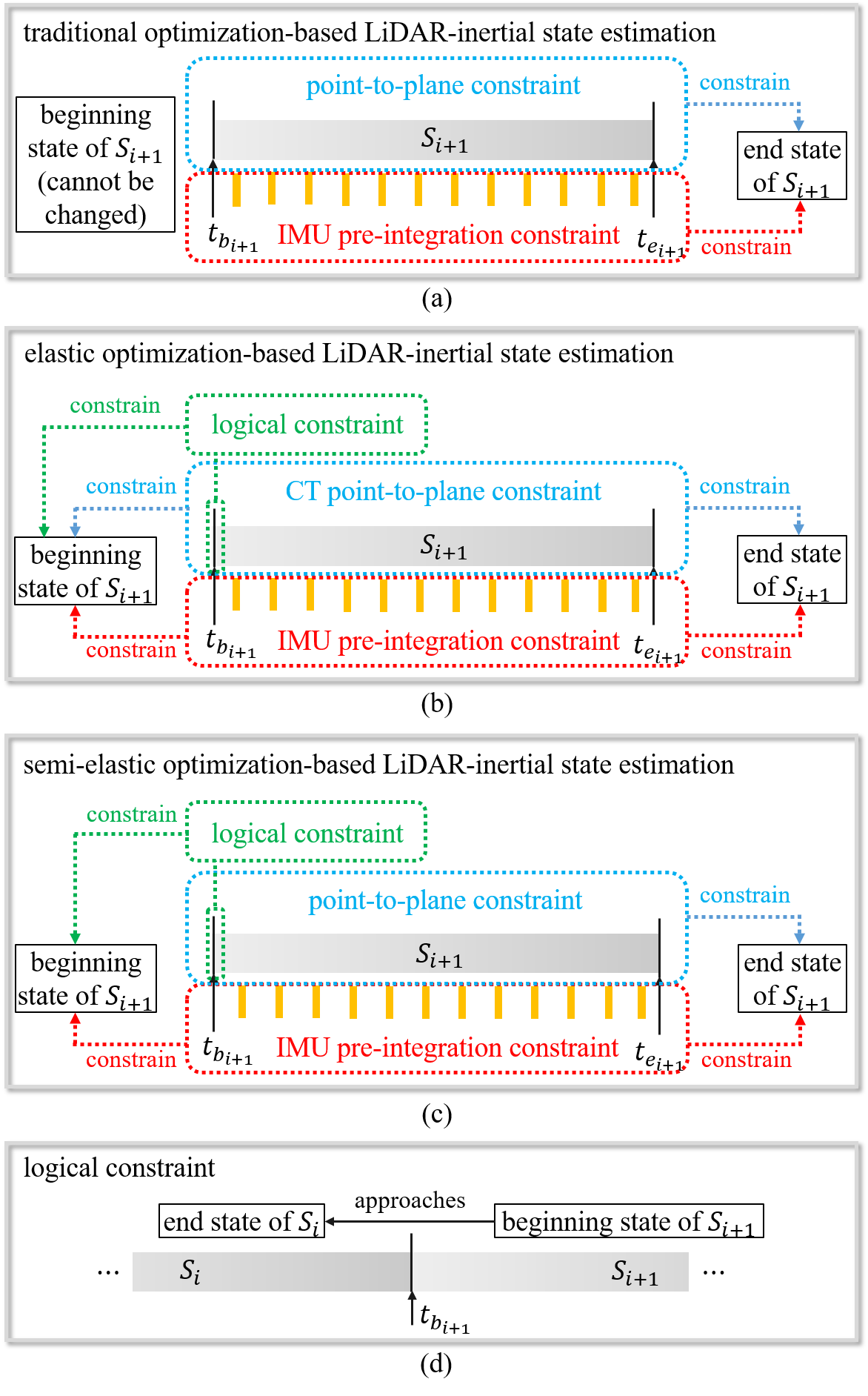}
		\caption{Illustration of (a) traditional optimization-based LiDAR-inertial state estimation, (b) elastic optimization-based LiDAR-inertial state estimation, (c) semi-elastic optimization-based LiDAR-inertial state estimation and (d) logical constraint. }
		\label{fig1}
	\end{center}
\end{figure}

As illustrated in Fig. \ref{fig1} (a), the traditional optimization-based LiDAR-inertial state estimation joints the traditional ICP and IMU pre-integration to constraint the state at the end of current sweep $S_{i+1}$ (i.e., the end state of $S_{i+1}$), while the state at the beginning of $S_{i+1}$ (i.e., the beginning state of $S_{i+1}$) is identical to the end state of $S_i$ and cannot be changed. Once the end state of $S_i$ is inaccurate, the error would entirely propagate to the end state of $S_{i+1}$ through IMU pre-integration constraint, yielding local inconsistency of the estimated state (e.g., zigzag trajectory or high-frequency oscillating velocity).

As illustrated in Fig. \ref{fig1} (b), the elastic ICP algorithm can be extended to elastic optimization-based LiDAR-inertial state estimation \cite{yuan2023liwo}, which utilizes continuous-time (CT) point-to-plane constraints \cite{dellenbach2022ct} and IMU pre-integration constraints \cite{qin2018vins} to optimize both the beginning and end state of $S_{i+1}$. Logically, the current sweep begins at the same time as the previous sweep ends. Therefore, a logical constraint (as illustrated in Fig. \ref{fig1} (d)) is employed to make the beginning state of $S_{i+1}$ approach the end state of $S_i$ during the state estimation. The logical constraint makes the beginning state of $S_{i+1}$ is no longer recklessly equal to the end state of $S_i$, but has sufficient elasticity to be adjusted. If the end state of $S_i$ wasn't solved accurately in last state estimation, the beginning state of $S_{i+1}$ has chance to be optimized to the correct value. By this way, the impact of last state accuracy on current state estimation is greatly reduced. However, it is important to note that the logical constraint may not always be strictly valid or stable. If the scene contains irregular geometry, such as trees, which could cause large CT point-to-plane errors, the weight of the logical constraint may decrease relative to CT point-to-plane constraints. When the logical constraint is not satisfied, both the beginning and end state of $S_i$ may become inaccurate and inconsistent. Conversely, if the weight of the logical constraint is set too high, the state's elastic space becomes limited in scenes with rich geometry, degenerate the elastic state estimation to the traditional state estimation. In conclusion, the logical constraint cannot work stably in the elastic state estimation framework, yielding poor robustness of the whole system.

In this paper, we propose a semi-elastic optimization-based LiDAR-inertial state estimation method for better balancing the logical constraint with other constraints, to ensure the local consistency and robustness of optimization-based LIO framework. As illustrated in Fig. \ref{fig1} (c), we employ the point-to-plane constraint to enhance the accuracy of the end state of $S_{i+1}$, and utilize a logical constraint to align the beginning state of $S_{i+1}$ to the end state of $S_i$. Furthermore, we impose IMU pre-integration constraint on both the beginning and end state of $S_{i+1}$ to ensure them consistently adhere to kinematic constraints within an elastic range. Unlike the scene-depending point-to-plane errors whose magnitude vary greatly with the scene, the magnitude of IMU pre-integration error largely depends on the sensor’s intrinsic noise parameters. This makes the weight relationship between logical constraint and IMU pre-integration constraint relatively valid and stable. As a result, the logical constraint remains valid and stable throughout the estimation process. We integrate our proposed semi-elastic optimization-based LiDAR-inertial state estimation into an optimization-based LIO framework. Experimental results on the public datasets $nclt$ \cite{carlevaris2016university}, $utbm$ \cite{yan2020eu}, $ulhk$ \cite{wen2020urbanloco}, $kaist$ \cite{jeong2019complex} demonstrate the following key findings: 1) Our system outperforms existing state-of-the-art LIO systems (i.e., \cite{li2021towards, shan2020lio, xu2022fast, chen2023direct}) in terms of a smaller absolute trajectory error (ATE); 2) The semi-elastic optimization-based LiDAR-inertial state estimation method can concurrently enhance accuracy, consistency, and robustness. A supplementary demo video showcases the compatibility of our system with Robosense LiDAR.

To summarize, this work includes one scientific contribution: We propose a novel semi-elastic optimization-based LiDAR-inertial state estimation method, which can better ensure the accuracy, consistency and robustness of the state estimation, and two technical contributions: 1) We develop a new LIO system based on our semi-elastic optimization-based state estimation method, and achieve the state-of-the-art accuracy; 2) We have released the source code of our system to benefit the development of the community.\footnote{https://github.com/ZikangYuan/semi\_elastic\_lio}

The rest of this paper is structured as follows. Sec. \ref{Related Work} reviews the relevant literature. Sec. \ref{Preliminary} provides preliminaries. Secs. \ref{System Overview} and \ref{System Details} presents system overview and details, followed by experimental evaluation in Sec. \ref{Experiments}. Sec. \ref{Conclusion} concludes the paper.

\section{Related Work}
\label{Related Work}

LIO-SAM \cite{shan2020lio} proposed an open-sourced optimization-based LIO framework, where the state solved by LiDAR ICP is the initial value of the node, and the IMU pre-integration constraint is as the edge. LINs \cite{qin2020lins} proposed an open-sourced error state iterated Kalman filter (ESIKF) based LIO framework. The state is estimated by balancing the weight of the LiDAR ICP constraint and the IMU constraint according to the iteratively updated Kalman gain. Fast-LIO \cite{xu2021fast} utilized the mathematical technique \cite{sorenson1966kalman} to optimize the process of solving Kalman gain, which converts the inverse calculation of the $N$-dimensional matrix to $18$-dimensional matrix, where $N$ is the number of point-to-plane residuals and $18$ is the dimensional of state vector. This transformation greatly speeds up the solving speed, and makes more point-to-plane residual terms can be added to LIO state estimation, thus improving the accuracy. Fast-LIO2 \cite{xu2022fast} proposed to build point-to-plane residuals directly on down-sampled input points. At the same time, the ikdtree \cite{cai2021ikd} is proposed to manage the map. Compared to the traditional kdtree, the ikdtree takes less time to traverse, add, and delete elements. DLIO \cite{chen2023direct} proposed to retain a 3-order minimum in state prediction and point distortion calibration to obtain more accurate state estimation results. SR-LIO \cite{yuan2022sr, yuan2024sr} adopt the sweep reconstruction method \cite{yuan2023sdv}, which segments and reconstructs raw input sweeps from spinning LiDAR to obtain reconstructed sweeps with higher frequency. Consequently, the frequency of estimated pose is also increased.

In addition to the above single sweep-to-map approaches, there are several LIO systems that use multi-sweep joint optimization framework. LIO\_mapping \cite{ye2019tightly} is an open-sourced LIO system based on optimization framework. It utilized a multi-sweep joint optimization framework similar to VINs-Mono \cite{qin2018vins} to ensure the accuracy of state estimation, however, the resulting large computational load causes LIO\_mapping cannot run in real time (but closer to real time). LiLi-OM \cite{li2021towards} selects some representative key-sweeps from original sweeps, and performs multi-sweep joint optimization to optimize the state of key-sweeps. However, the real-time performance cannot be guaranteed steadily. LIO\_Livox \cite{livox2021} is an official open-sourced LIO framework proposed by Livox company. For the point cloud of input sweep, the dynamic points are removed, and the edge, surface and irregular features are extracted. Then the state is estimated by multi-sweep joint optimization. However, the official LIO\_Livox cannot ensure the real-time performance if utilizing more than 3 sweeps to perform multi-sweep joint LIO optimization. Ssl\_slam3 \cite{wang2021lightweight} utilized the point-to-plane constraints from of a single sweep and IMU pre-integration constraints from multiple sweeps to improve real-time performance. All of above systems utilized multi-sweep joint optimization to perform LIO state estimation. In addition to eliminating accumulated errors, multi-sweep joint optimization gives sufficient elasticity to state, making them more likely to update to the correct value. However, introducing more variables and residuals will also increase the amount of computation, so the side effect is also very significant. By contrast, our semi-elastic optimization-based LIO optimization framework avoids additional computation while providing enough elasticity for state variables.

\section{Preliminary}
\label{Preliminary}

\subsection{Coordinate Systems}
\label{Coordinate Systems}

We denote $(\cdot)^w$, $(\cdot)^l$ and $(\cdot)^o$ as a 3D point in the world coordinates, the LiDAR coordinates and the IMU coordinates respectively. The world coordinate is coinciding with $(\cdot)^o$ at the starting position.

We denote the LiDAR coordinate for taking the $i_{th}$ sweep at time $t_i$ as $l_i$ and the corresponding IMU coordinate at $t_i$ as $o_i$, then the transformation matrix (i.e., external parameters) from $l_i$ to $o_i$ is denoted as $\mathbf{T}_{l_i}^{o_i} \in S E(3)$, which consists of a rotation matrix $\mathbf{R}_{l_i}^{o_i} \in S O(3)$ and a translation vector $\mathbf{t}_{l_i}^{o_i} \in \mathbb{R}^3$. The external parameters are usually calibrated once offline and remain constants during online state estimation; therefore, we can represent $\mathbf{T}_{l_i}^{o_i}$ using $\mathbf{T}_{l}^{o}$ for simplicity. In the following statement, we omit the index that represents the coordinate system for simplified notation. For instance, the pose from the IMU coordinate to the world coordinate is strictly defined as $\mathbf{T}_{o_i}^{w}$. Now we define it as $\mathbf{T}_{i}^{w}$ for simplify.

In addition to pose, we also estimate the velocity $\mathbf{v}$, the accelerometer bias ${\mathbf{b}}_{\mathbf{a}}$ and the gyroscope bias ${\mathbf{b}}_{\boldsymbol{\omega}}$, which are represented by a state vector:
\begin{equation}
	\label{equation2}
	\boldsymbol{x}=\left[\mathbf{t}^T, \mathbf{q}^T, \mathbf{v}^T, \mathbf{b}_{\mathbf{a}}{ }^T, \mathbf{b}_{\boldsymbol{\omega}}{ }^T\right]^T
\end{equation}
where $\mathbf{q}$ is the quaternion form of the rotation matrix $\mathbf{R}$.

\subsection{IMU Measurement Model}
\label{IMU Measurement Model}

A IMU consists of an accelerometer and a gyroscope. The raw accelerometer and gyroscope measurements from IMU, $\hat{\mathbf{a}}_t$ and $\hat{\boldsymbol{\omega}}_t$, are given by:
\begin{equation}
	\label{equation3}
	\begin{gathered}
		\hat{\mathbf{a}}_t=\mathbf{a}_t+\mathbf{b}_{\mathbf{a}_t}+\mathbf{R}_w^t \mathbf{g}^w+\mathbf{n}_{\mathbf{a}} \\
		\hat{\boldsymbol{\omega}}_t=\boldsymbol{\omega}_t+\mathbf{b}_{\boldsymbol{\omega}_t}+\mathbf{n}_{\boldsymbol{\omega}}
	\end{gathered}
\end{equation}
IMU measurements, which are measured in IMU coordinates, combine the force for countering gravity and the platform dynamics, and are affected by acceleration bias $\mathbf{b}_{\mathbf{a}_t}$, gyroscope bias $\mathbf{b}_{\boldsymbol{\omega}_t}$, and additive noise. As mentioned in VINs-Mono \cite{qin2018vins}, the additive noise in acceleration and gyroscope measurements are modeled as Gaussian white noise, $\mathbf{n}_{\mathbf{a}} \sim N\left(\mathbf{0}, \boldsymbol{\sigma}_{\mathbf{a}}^2\right)$, $\mathbf{n}_{\boldsymbol{\omega}} \sim N\left(\mathbf{0}, \boldsymbol{\sigma}_{\boldsymbol{\omega}}^2\right)$. Acceleration bias and gyroscope bias are modeled as random walk, whose derivatives are Gaussian, $\dot{\mathbf{b}}_{\mathbf{a}_t}=\mathbf{n}_{\mathbf{b}_{\mathbf{a}}} \sim N\left(\mathbf{0}, \boldsymbol{\sigma}_{\mathbf{b}_{\mathbf{a}}}^2\right)$, $\dot{\mathbf{b}}_{{\boldsymbol{\omega}}_t}=\mathbf{n}_{\mathbf{b}_{\boldsymbol{\omega}}} \sim N\left(\mathbf{0}, \boldsymbol{\sigma}_{\mathbf{b}_{\boldsymbol{\omega}}}^2\right)$.

\subsection{Sweep State Expression}
\label{Sweep State Expression}

The same as CT-ICP \cite{dellenbach2022ct}, we represent the state of a sweep $S$ by: 1) the state at the beginning time $t_b$ of $S$ (e.g., $\boldsymbol{x}_b$) and 2) the state at the end time $t_e$ of $S$ (e.g., $\boldsymbol{x}_e$). There are two ways for LiDAR point distortion calibration in our system: utilizing IMU integrated pose to calibrate or utilizing uniform motion model. For vehicle platforms that are mostly in uniform motion state, we recommend the uniform motion distortion calibration model because IMU measurements are sometimes affected by large noise.

\section{System Overview}
\label{System Overview}

\begin{figure*}
	\begin{center}
		\includegraphics[scale=0.51]{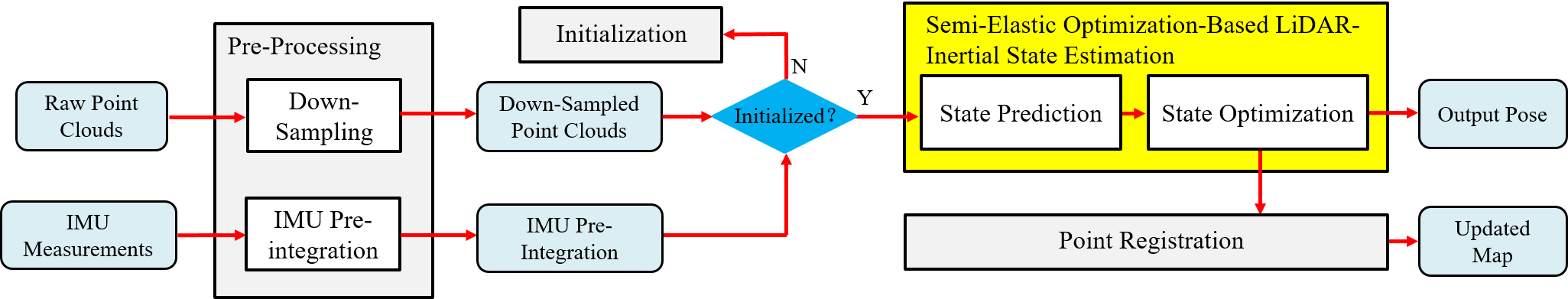}
		\caption{Overview of our LIO system which consists of four main modules: a pre-processing module, an initialization module, a state estimation module and a point registration module. The yellow part is the semi-elastic optimization-based LiDAR-inertial state estimation we proposed.}
		\label{fig2}
	\end{center}
\end{figure*}

Fig. \ref{fig2} illustrates the framework of our system which consists of four main modules: a pre-processing module, an initialization module, a semi-elastic optimization-based LiDAR-inertial state estimation module and a point registration module. The pre-processing module takes sensor raw data as input, and outputs down-sampled point clouds and IMU pre-integration. The initialization module is used to estimate some state parameters such as gravitational acceleration, accelerometer bias, gyroscope bias, and initial velocity. The semi-elastic optimization-based LiDAR-inertial state estimation module executes state prediction and state optimization in turn, which is detailed in Sec. \ref{Semi-Elastic LiDAR-Inertial State Estimation}. After state estimation, the point registration adds the new points to the map and delete the points that are far away from current position.

\section{System Details}
\label{System Details}

\subsection{Pre-processing}
\label{Pre-processing}

\subsubsection{Down-Sampling}
\label{Down-Sampling}

Due to huge number of 3D point clouds to be processed, the computational burden of the whole system is heavy. In order to reduce the computational burden, we down-sample the input point clouds. Firstly, we perform the quantitative down-sampling strategy, which keeps only one out of every four points. Then, we put the quantitative down-sampled points into a volume with $0.5\times0.5\times0.5$ (unit: m) voxel size, and make each voxel contain only one point, which is the same as CT-ICP \cite{dellenbach2022ct}.

\subsubsection{IMU Pre-integration}
\label{IMU Pre-integration}

Typically, the IMU sends out data at a much higher frequency than the LiDAR. Pre-integration of all IMU measurements between two consecutive sweeps $S_i$ and $S_{i+1}$ can well summarize the dynamics of the hardware platform from time $t_{e_i}$ to $t_{e_{i+1}}$, where $e_i$ and $e_{i+1}$ are the end time stamp of $S_i$ and $S_{i+1}$ respectively. In this work, we employ the discrete-time quaternion-based derivation of IMU pre-integration approach \cite{lupton2011visual}, and incorporate IMU bias using the method in \cite{qin2018vins}. Specifically, the pre-integrations between $S_i$ and $S_{i+1}$ in the corresponding IMU coordinates $o_{e_i}$ and $o_{e_{i+1}}$, i.e., $\hat{\boldsymbol{\alpha}}_{e_{i+1}}^{e_i}$, $\hat{\boldsymbol{\beta}}_{e_{i+1}}^{e_i}$, and $\hat{\boldsymbol{\gamma}}_{e_{i+1}}^{e_i}$, are calculated, where $\boldsymbol{\alpha}_{e_{i+1}}^{e_i}$, $\boldsymbol{\beta}_{e_{i+1}}^{e_i}$, $\boldsymbol{\gamma}_{e_{i+1}}^{e_i}$ are the pre-integration of translation, velocity, rotation from IMU measurements respectively. In addition, the Jacobian of pre-integration with respect to bias, i.e., $\mathbf{J}_{\mathbf{b}_{\mathbf{a}}}^{\boldsymbol{\alpha}}$, $\mathbf{J}_{\mathbf{b}_{\boldsymbol{\omega}}}^{\boldsymbol{\alpha}}$, $\mathbf{J}_{\mathbf{b}_{\mathbf{a}}}^{\boldsymbol{\beta}}$, $\mathbf{J}_{\mathbf{b}_{\boldsymbol{\omega}}}^{\boldsymbol{\beta}}$, $\mathbf{J}_{\mathbf{b}_{\boldsymbol{\omega}}}^{\boldsymbol{\gamma}}$, are also calculated according to the error state kinematics.

\subsection{Initialization}
\label{Initialization}

The initialization module aims to estimate all necessary values including initial pose, velocity, gravitational acceleration, accelerometer bias and gyroscope bias, for subsequent state estimation. We adopt static initialization in our system. Please refer to \cite{geneva2020openvins} for more details about our initialization module.

\subsection{Semi-Elastic LiDAR-Inertial State Estimation}
\label{Semi-Elastic LiDAR-Inertial State Estimation}

\subsubsection{State Prediction}
\label{State Prediction}

When every new down-sampled sweep $P_{i+1}$ completes, we use IMU measurements to predict the state at the beginning of $P_{i+1}$ (i.e., $\boldsymbol{x}_{b_{i+1}}^w$) and the state at the end of $P_{i+1}$ (i.e., $\boldsymbol{x}_{e_{i+1}}^w$), which are provided as the prior motion for LIO-optimization. Specifically, the predicted state $\boldsymbol{x}_{b_{i+1}}^w$ (i.e., $\mathbf{t}_{b_{i+1}}^w$, $\mathbf{R}_{b_{i+1}}^w$, $\mathbf{v}_{b_{i+1}}^w$, ${\mathbf{b}}_{\mathbf{a}_{b_{i+1}}}$ and ${\mathbf{b}}_{\boldsymbol{\omega}_{b_{i+1}}}$) is assigned as:
\begin{equation}
	\label{equation4}
	\boldsymbol{x}_{b_{i+1}}^w = \boldsymbol{x}_{e_{i}}^w
\end{equation}
and $\boldsymbol{x}_{e_{i+1}}^w$ (i.e., $\mathbf{t}_{e_{i+1}}^w$, $\mathbf{R}_{e_{i+1}}^w$, $\mathbf{v}_{e_{i+1}}^w$, ${\mathbf{b}}_{\mathbf{a}_{e_{i+1}}}$ and ${\mathbf{b}}_{\boldsymbol{\omega}_{e_{i+1}}}$) is calculated as:
\begin{equation}
	\label{equation5}
	\begin{gathered}
		\mathbf{R}_{n+1}^w=\mathbf{R}_n^w Exp\left(\left(\frac{\hat{\boldsymbol{\omega}}_n+\hat{\boldsymbol{\omega}}_{n+1}}{2}-\mathbf{b}_{\boldsymbol{\omega}_{e_i}}\right) \delta t\right) \\
		\mathbf{v}_{n+1}^w=\mathbf{v}_n^w+\mathbf{R}_n^w\left(\frac{\hat{\mathbf{a}}_n+\hat{\mathbf{a}}_{n+1}}{2}-\mathbf{b}_{\mathbf{a}_{e_i}}-\mathbf{R}_w^n \mathbf{g}^w\right) \delta t \\
		\mathbf{t}_{n+1}^w=\mathbf{t}_n^w+\mathbf{v}_n^w \delta t+ \\ \frac{1}{2}\mathbf{R}_n^w\left(\frac{\hat{\mathbf{a}}_n+\hat{\mathbf{a}}_{n+1}}{2}-\mathbf{b}_{\mathbf{a}_{e_i}}-\mathbf{R}_w^n \mathbf{g}^w\right) \delta t^2
	\end{gathered}
\end{equation}
where $\mathbf{g}^w$ is the gravitational acceleration in world coordinates, $n$ and $n+1$ are two time instants of obtaining an IMU measurement during $\left[t_{e_i}, t_{e_{i+1}}\right]$, $\delta t$ is the time interval between $n$ and $n+1$. We iteratively increase $n$ from $0$ to $\left(t_{e_{i+1}}-t_{e_i}\right) / \delta t$ to obtain $\boldsymbol{x}_{e_{i+1}}^w$. When $n=0$, $\boldsymbol{x}_{n}^w=\boldsymbol{x}_{e_i}^w$. For $\mathbf{b}_{\mathbf{a}_{e_{i+1}}}$ and $\mathbf{b}_{\boldsymbol{\omega}_{e_{i+1}}}$, we set the predicted values of them by: $\mathbf{b}_{\mathbf{a}_{e_{i+1}}}=\mathbf{b}_{\mathbf{a}_{e_i}}$ and $\mathbf{b}_{\boldsymbol{\omega}_{e_{i+1}}}=\mathbf{b}_{\boldsymbol{\omega}_{e_i}}$.

\subsubsection{State Optimization}
\label{State Optimization}

We jointly utilize measurements of the LiDAR and IMU to optimize the beginning state (i.e., $\boldsymbol{x}_{b_{i+1}}^w$) and the end state (i.e., $\boldsymbol{x}_{e_{i+1}}^w$) of the current sweep $P_{i+1}$, where the variable vector is expressed as:
\begin{equation}
	\label{equation6}
	\boldsymbol{\chi}=\left\{\boldsymbol{x}_{b_{i+1}}^w, \boldsymbol{x}_{e_{i+1}}^w\right\}
\end{equation}
\textbf{Residual from the LiDAR constraint.} For a distortion-calibrated point $\mathbf{p}$, we first project $\mathbf{p}$ to the world coordinate to obtain $\mathbf{p}^w$, and then find 20 nearest points around $\mathbf{p}^w$ from the volume. To search for the nearest neighbor of $\mathbf{p}^w$. We only search in the voxel $V$ to which $\mathbf{p}^w$ belongs, and the 8 voxels adjacent to $V$. The 20 nearest points are used to fit a plane with a normal $\mathbf{n}$ and a distance $d$. Accordingly, we can build the point-to-plane residual $r^{\mathbf{p}}$ for $\mathbf{p}$ as:
\begin{equation}
	\label{equation7}
	\begin{gathered}
		r^{\mathbf{p}}=\omega_{\mathbf{p}}\left(\mathbf{n}^T \mathbf{p}^w+d\right) \\
		\mathbf{p}^w=\mathbf{q}_{e_{i+1}}^w \mathbf{p}+\mathbf{t}_{e_{i+1}}^w
	\end{gathered}
\end{equation}
where $\omega_{\mathbf{p}}$ is a weight parameter utilized in \cite{dellenbach2022ct}, $\mathbf{q}_{e_{i+1}}^w$ is the rotation with respect to $(\cdot)^w$ at $t_{e_{i+1}}$, $\mathbf{t}_{e_{i+1}}^w$ is the translation with respect to $(\cdot)^w$ at $t_{e_{i+1}}$. Both $\mathbf{q}_{e_{i+1}}^w$, $\mathbf{t}_{e_{i+1}}^w$ are variables to be refined, and the initial value of them are obtained from Sec. \ref{State Prediction}.

\textbf{Residual from the IMU constraint.} Considering the IMU measurements during $\left[t_{e_i}, t_{e_{i+1}}\right]$, according to pre-integration introduced in Sec. \ref{IMU Pre-integration}, the residual for pre-integrated IMU measurements can be computed as:
\begin{equation}
	\label{equation8}
	\begin{gathered}
		{{\mathbf{r}_o}_{{e_{i+1}}}^{b_{i+1}}}= \\
		{\left[\begin{array}{c}
				\mathbf{R}_w^{b_{i+1}}\left(\mathbf{t}_{e_{i+1}}^w-\mathbf{t}_{b_{i+1}}^w+\frac{1}{2} \mathbf{g}^w \Delta t^2-\mathbf{v}_{b_{i+1}}^w \Delta t\right)-\hat{\boldsymbol{\alpha}}_{e_{i+1}}^{e_i} \\
				\mathbf{R}_w^{b_{i+1}}\left(\mathbf{v}_{e_{i+1}}^w+\mathbf{g}^w \Delta t-\mathbf{v}_{b_{i+1}}^w\right)-\hat{\boldsymbol{\beta}}_{e_{i+1}}^{e_i} \\
				2\left[\mathbf{q}_{b_{i+1}}^{{w}^{-1}} \otimes \mathbf{q}_{e_{i+1}}^w \otimes\left(\hat{\boldsymbol{\gamma}}_{e_{i+1}}^{e_i}\right)^{-1}\right]_{x y z} \\
				\mathbf{b}_{\mathbf{a}_{i+1}}-\mathbf{b}_{\mathbf{a}_i} \\
				\mathbf{b}_{\boldsymbol{\omega}_{i+1}}-\mathbf{b}_{\boldsymbol{\omega}_i}
			\end{array}\right]}
	\end{gathered}
\end{equation}
where $[\cdot]_{x y z}$ extracts the vector part of a quaternion q for error state representation. At the end of each iteration, we update $\left[\hat{\boldsymbol{\alpha}}_{e_{i+1}}^{e_i}, \hat{\boldsymbol{\beta}}_{e_{i+1}}^{e_i}, \hat{\boldsymbol{\gamma}}_{e_{i+1}}^{e_i}\right]^T$ with the first order Jacobian approximation \cite{qin2018vins}.

\textbf{Residual from the logical constraint.} $\boldsymbol{x}_{b_{i+1}}^w$ and $\boldsymbol{x}_{e_i}^w$ are two states at the same time stamp $t_{b_{i+1}}$($t_{e_i}$). Logically, $\boldsymbol{x}_{e_i}^w$ and $\boldsymbol{x}_{b_{i+1}}^w$ should be the same. Therefore, we build the logical consistency residual as follow:
\begin{equation}
	\label{equation9}
	\mathbf{r}_c=\left[\begin{array}{c}
		\mathbf{r}_c^{\mathbf{t}} \\
		\mathbf{r}_c^{\mathbf{q}} \\
		\mathbf{r}_c^{\mathbf{v}} \\
		\mathbf{r}_c^{\mathbf{b}_{\mathbf{a}}} \\
		\mathbf{r}_c^{\mathbf{b_{\boldsymbol{\omega}}}}
	\end{array}\right]=\left[\begin{array}{c}
		\mathbf{t}_{b_{i+1}}^w-\mathbf{t}_{e_i}^w \\
		2\left[{\mathbf{q}_{e_i}^{w}}^{-1} \otimes \mathbf{q}_{b_{i+1}}^w\right]_{x y z} \\
		\mathbf{v}_{b_{i+1}}^w-\mathbf{v}_{e_i}^w \\
		\mathbf{b}_{\mathbf{a}_{b_{i+1}}}-\mathbf{b}_{\mathbf{a}_{e_i}} \\
		\mathbf{b}_{\boldsymbol{\omega}_{b_{i+1}}}-\mathbf{b}_{\boldsymbol{\omega}_{e_i}}
	\end{array}\right]
\end{equation}
where $\mathbf{t}_{b_{i+1}}^w$, $\mathbf{q}_{b_{i+1}}^w$, $\mathbf{v}_{b_{i+1}}^w$, $\mathbf{b}_{\mathbf{a}_{b_{i+1}}}$, $\mathbf{b}_{\boldsymbol{\omega}_{b_{i+1}}}$ are varibales to be optimized.

By minimizing the Mahanalobis distance of point-to-plane residuals, the IMU pre-integration residuals, and the logical consistency residuals, we maximize a posteriori estimation as:
\begin{equation}
	\label{equation10}
	\begin{gathered}
		\boldsymbol{\chi}=\min _{\boldsymbol{\chi}} \\ \left\{\rho\left(\sum_{\mathbf{p} \in P_{i+1}}\left\|r^{\mathbf{p}}\right\|_{\mathbf{P}_L}^2+\left\|{\mathbf{r}_o}_{e_{i+1}}^{b_{i+1}}\right\|_{\mathbf{P}_{e_{i+1}}^{e_i}}^2+\left\|\mathbf{r}_c\right\|^2\right)\right\}
	\end{gathered}
\end{equation}
where $\rho$ is the Huber kernel to eliminate the influence of outlier residuals. $\mathbf{P}_{e_{i+1}}^{e_i}$ is the covariance matrix of pre-integrated IMU measurements. The inverse of $\mathbf{P}_{e_{i+1}}^{e_i}$ is utilized as the weight of IMU pre-integration residuals. $\mathbf{P}_L$ is a diagonal matrix while each element is a constant (e.g., 0.001 in our system) to indicate the reliability of the point-to-plane residuals. The inverse of $\mathbf{P}_L$ is utilized as the weight of point-to-plane residuals. After finishing the state optimization, we selectively add the points of current sweep to the map.

\subsection{Point Registration}
\label{Point Registration}

Following CT-ICP \cite{dellenbach2022ct}, the cloud map is stored in a volume, and the size of each voxel is 1.0$\times$1.0$\times$1.0 (unit: m). Each voxel contains a maximum of 20 points. When the state of current down-sampled sweep $P_{i+1}$ has been estimated, we transform $P_{i+1}$ to the world coordinate system $(\cdot)^w$, and add the transformed points into the volume map. If a voxel already has 20 points, the new points cannot be added to it. Meanwhile, we delete the points that are far away from current position.

\section{Experiments}
\label{Experiments}

\begin{table}[]
	\begin{center}
		\caption{Datasets of All Sequences for Evaluation}
		\label{table2}
		\begin{tabular}{cccc}
			\hline
			& Name         & \begin{tabular}[c]{@{}c@{}}Duration\\ (min:sec)\end{tabular} & \begin{tabular}[c]{@{}c@{}}Distance\\ (km)\end{tabular} \\ \hline
			$nclt\_1$  & 2012-01-08   & 92:16                                                        & 6.4                                                     \\
			$nclt\_2$  & 2012-01-22   & 86:11                                                        & 6.1                                                     \\
			$nclt\_3$  & 2012-02-04   & 77:39                                                        & 5.5                                                     \\
			$nclt\_4$  & 2012-02-05   & 93:40                                                        & 6.5                                                     \\
			$nclt\_5$  & 2012-02-12   & 85:17                                                        & 5.8                                                     \\
			$nclt\_6$  & 2012-03-17   & 81:51                                                        & 5.8                                                     \\
			$nclt\_7$  & 2012-04-29   & 43:17                                                        & 3.1                                                     \\
			$nclt\_8$  & 2012-05-11   & 83:36                                                        & 6.0                                                     \\
			$nclt\_9$  & 2012-05-26   & 97:23                                                        & 6.3                                                     \\
			$nclt\_10$ & 2012-06-15   & 55:10                                                        & 4.1                                                     \\
			$nclt\_11$ & 2012-08-04   & 79:27                                                        & 5.5                                                     \\
			$nclt\_12$ & 2012-08-20   & 88:44                                                        & 6.0                                                     \\
			$nclt\_13$ & 2012-09-28   & 76:40                                                        & 5.6                                                     \\
			$nclt\_14$ & 2012-11-04   & 79:53                                                        & 4.8                                                     \\
			$nclt\_15$ & 2012-12-01   & 75:50                                                        & 5.0                                                     \\
			$nclt\_16$ & 2013-01-10   & 17:02                                                        & 1.1                                                     \\
			$utbm\_1$  & 2018-07-19   & 15:26                                                        & 4.98                                                    \\
			$utbm\_2$  & 2019-01-31   & 16:00                                                        & 6.40                                                    \\
			$utbm\_3$  & 2019-04-18   & 11:59                                                        & 5.11                                                    \\
			$utbm\_4$  & 2018-07-20   & 16:45                                                        & 4.99                                                    \\
			$utbm\_5$  & 2018-07-17   & 15:59                                                        & 4.99                                                    \\
			$utbm\_6$  & 2018-07-16   & 15:59                                                        & 4.99                                                    \\
			$utbm\_7$  & 2018-07-13   & 16:59                                                        & 5.03                                                    \\
			$ulhk\_1$  & 2019-01-17   & 5:18                                                         & 0.60                                                    \\
			$ulhk\_2$  & 2019-04-26-1 & 2:30                                                         & 0.55                                                    \\
			$kaist\_1$ & urban\_07    & 9:16                                                         & 2.55                                                    \\
			$kaist\_2$ & urban\_08    & 5:07                                                         & 1.56                                                    \\
			$kaist\_3$ & urban\_13    & 24:14                                                        & 2.36                                                    \\ \hline
		\end{tabular}
	\end{center}
\end{table}

We evaluated our system on the public datasets $nclt$ \cite{carlevaris2016university}, $utbm$ \cite{yan2020eu}, $ulhk$ \cite{wen2020urbanloco} and $kaist$ \cite{jeong2019complex}. $nclt$ is a large-scale, long-term autonomy unmanned ground vehicle dataset collected in the University of Michigans North Campus. The $nclt$ dataset contains a full data stream from a Velodyne HDL-32E LiDAR and 50\,Hz data from Microstrain MS25 IMU. The $nclt$ dataset has a much longer duration and amount of data than other datasets and contains several open scenes, such as a large open parking lot. Different from the other three datasets (e.g., $utbm$, $ulhk$ and $kaist$), the LiDAR of $nclt$ need 130$\sim$140\,ms to finish a 360\,deg sweep, which means the frequency of sweep is around 7.5\,Hz. In addition, 50\,Hz IMU measurements cannot meet the requirements of some systems (e.g., LIO-SAM \cite{shan2020lio}). Therefore, we increase the frequency of the IMU to 100\,Hz by interpolation. The $utbm$ dataset is collected with a human-driving robocar in maximum 50\,km/h speed, which has two 10\,Hz Velodyne HDL-32E LiDAR5 and 100\,Hz Xsens MTi-28A53G25 IMU. For point clouds, we only utilize the data from the left LiDAR. $ulhk$ contains 10\,Hz LiDAR sweep from Velodyne HDL-32E and 100\,Hz IMU data from a 9-axis Xsens MTi-10 IMU. $kaist$ contains two 10\,Hz Velodyne VLP-16, 200\,Hz Ssens MTi-300 IMU and 100\,Hz RLS LM13 wheel encoder. Two 3D LiDARs are tilted by approximately $45^{\circ}$. For point clouds, we utilize the data from both two 3D LiDARs. All the sequences of $utbm$, $ulhk$ and $kaist$ are collected in structured urban areas by a human-driving vehicle. Due to the above four datasets both utilize the vehicle platform, we utilize uniform motion distortion calibration model in our system. The details about all the 28 sequences used in this section, including name, duration, and distance, are listed in Table \ref{table2}. For all four datasets, we utilize the absolute translational error (ATE) as the evaluation metrics. A consumer-level computer equipped with an Intel Core i7-12700 and 32 GB RAM is used for all experiments.

\subsection{Comparison of the State-of-the-Arts}
\label{Comparison of the State-of-the-Arts}

\begin{table}[]
	\begin{center}
		\caption{RMSE of ATE Comparison of State-of-the-art (Unit: m)}
		\label{table3}
		\begin{threeparttable}
			\begin{tabular}{c|cccc|c}
				\hline
				& LiLi-OM & LIO-SAM & Fast-LIO2 & DLIO           & Ours           \\ \hline
				$nclt\_1$  & 50.71                                              & 1.85*                                              & 3.57                                                 & 3.27           & \textbf{1.76}  \\
				$nclt\_2$  & 91.20                                              & 9.70*                                              & 2.24                                                 & 2.82           & \textbf{1.82}  \\
				$nclt\_3$  & 92.93                                              & 2.16*                                              & 2.77                                                 & 5.35*          & \textbf{1.85}  \\
				$nclt\_4$  & 215.91                                             & 2.70*                                              & 3.60                                                 & 18.10*         & \textbf{1.55}  \\
				$nclt\_5$  & 145.52                                             & 2.74*                                              & 5.23                                                 & 2.01*          & \textbf{1.82}  \\
				$nclt\_6$  & 262.30                                             & $\times$                                                  & 3.04                                                 & 6.39           & \textbf{2.17}  \\
				$nclt\_7$  & 93.61                                              & 1.75                                               & 1.40                                                 & 1.51           & \textbf{1.38}  \\
				$nclt\_8$  & 185.24                                             & $\times$                                                  & 2.46                                                 & 3.14           & \textbf{2.20}  \\
				$nclt\_9$  & 141.83                                             & $\times$                                                  & 2.60                                                 & 12.44          & \textbf{2.22}  \\
				$nclt\_10$ & 50.42                                              & 2.97                                               & 2.37                                                 & 2.98           & \textbf{2.24}  \\
				$nclt\_11$ & 137.05                                             & 2.26*                                              & 2.59                                                 & 7.84           & \textbf{2.25}  \\
				$nclt\_12$ & 224.68                                             & 10.68*                                             & 4.01                                                 & 2.46           & \textbf{2.23}  \\
				$nclt\_13$ & $\times$                                                  & $\times$                                                  & 2.65                                                 & 7.72           & \textbf{2.13}  \\
				$nclt\_14$ & 229.66                                             & 7.89*                                              & 3.83                                                 & 3.15           & \textbf{2.19}  \\
				$nclt\_15$ & $\times$                                                  & $\times$                                                  & 4.37                                                 & 3.89           & \textbf{2.16}  \\
				$nclt\_16$ & $\times$                                                  & 1.78                                               & \textbf{0.90}                                        & 0.91           & 0.92           \\
				$utbm\_1$  & 67.16                                              & -                                                  & 15.13                                                & \textbf{14.25} & 15.27          \\
				$utbm\_2$  & 38.17                                              & -                                                  & 21.21                                                & \textbf{13.85} & 18.58          \\
				$utbm\_3$  & 10.70                                              & -                                                  & 10.81                                                & 55.28          & \textbf{10.16} \\
				$utbm\_4$  & 70.98                                              & -                                                  & 15.20                                                & 18.05          & \textbf{13.44} \\
				$utbm\_5$  & 86.80                                              & -                                                  & 13.16                                                & 12.66          & \textbf{11.71} \\
				$utbm\_6$  & 84.77                                              & -                                                  & 14.67                                                & 13.42 & \textbf{13.40}          \\
				$utbm\_7$  & 62.57                                              & -                                                  & 13.24                                                & 14.95          & \textbf{12.15} \\
				$ulhk\_1$  & $\times$                                                  & 1.68                                               & 1.20                                                 & 2.44           & \textbf{0.93}  \\
				$ulhk\_2$  & \textbf{3.11}                                      & 3.13                                               & 3.24                                                 & $\times$              & 3.24           \\
				$kaist\_1$ & $\times$                                                  & 16.96                                              & 0.88                                                 & 1.04           & \textbf{0.86}  \\
				$kaist\_2$ & $\times$                                                  & $\times$                                                  & 16.27                                                & \textbf{1.91}  & 3.44           \\
				$kaist\_3$ & $\times$                                                  & $\times$                                                  & $\times$                                                    & $\times$              & \textbf{1.04}  \\ \hline
			\end{tabular}
		\end{threeparttable}
		\begin{tablenotes}
			\footnotesize
			\item[] \textbf{Denotations}: “$\times$” means the system fails to run entirety on the corresponding sequence, “-” means the corresponding value is not available, and “*” means the system crashes towards the end of this sequence.
		\end{tablenotes}
	\end{center}
\end{table}

We compare our system with four state-of-the-art LIO systems, i.e., LiLi-OM \cite{li2021towards}, LIO-SAM \cite{shan2020lio}, Fast-LIO2 \cite{xu2022fast} and DLIO \cite{chen2023direct}. LiLi-OM and LIO-SAM joints LiDAR point-to-plane residuals and IMU pre-integration residuals into an optimization-based framework, where LiLi-OM selects key sweeps from raw measurements and LIO-SAM processes each input sweep. Both Fast-LIO2 and DLIO are ESIKF-based LIO systems, where Fast-LIO2 utilizes the ikdtree \cite{cai2021ikd} and DLIO utilizes the nanoflan \cite{blanco2014nanoflann} to manage the map. Both LiLi-OM and LIO-SAM are based on the traditional optimization-based LiDAR-inertial state estimation framework. Fast-LIO2 and DLIO are ESIKF-based LIO systems, and the concepts of "traditional", "elastic" and "semi-elastic" proposed in this paper are for the optimization-based framework, so they don't have a strict classification. For a fair comparison, we obtain the results of above systems based on the source code provided by the authors.

Results in Table \ref{table3} demonstrate that our system outperforms LiLi-OM, LIO-SAM and Fast-LIO2 for almost all sequences in terms of smaller ATE. Although our accuracy is not the best on $nclt\_16$ and $ulhk\_2$, we are very close to the best accuracy, where the ATE is only 2\,cm more on $nclt\_16$ and 13\,cm more on $ulhk\_4$. “-” means the corresponding value is not available. LIO-SAM needs 9-axis IMU data as input, while the $utbm$ dataset only provides 6-axis IMU data. Therefore, we cannot provide the results of LIO-SAM on $utbm$ dataset. “$\times$” means the system fails to run entirety on the corresponding sequence, and “*” means the system crashes towards the end of this sequence. Although DLIO achieves more accurate results than our system on some sequences (i.e., $utbm\_1$, $utbm\_2$ and $kaist\_2$), it cannot run completely on more sequences or can run but achieve a huge ATE error (i.e., $nclt\_3$, $nclt\_4$, $nclt\_5$, $nclt\_9$, $utbm\_3$, $ulhk\_2$ and $kaist\_3$). LiLi-OM and LIO-SAM fails or crashes on most sequences, and the state-of-the-art Fast-LIO2 also fails on $kaist\_3$ and achieves a huge ATE error on $kaist\_2$. In contrast, our system can run properly on almost all sequences, which also demonstrates the good robustness of our system.

\subsection{Ablation Study of Semi-Elastic State Estimation}
\label{Ablation Study of Semi-Elastic State Estimation}

\begin{table}[]
	\begin{center}
		\caption{Ablation Study of Semi-Elastic State Estimation on Accuracy}
		\label{table4}
		\begin{threeparttable}
			\begin{tabular}{p{1.55cm}<{\centering}|p{1.55cm}<{\centering}p{1.55cm}<{\centering}|p{1.55cm}<{\centering}}
				\hline
				& traditional   & elastic        & semi-elastic   \\ \hline
				$nclt\_1$  & 2.74          & $\times$              & \textbf{1.76}  \\
				$nclt\_2$  & $\times$             & $\times$              & \textbf{1.82}  \\
				$nclt\_3$  & 5.12          & $\times$              & \textbf{1.85}  \\
				$nclt\_4$  & 2.05          & $\times$              & \textbf{1.55}  \\
				$nclt\_5$  & 2.05          & $\times$              & \textbf{1.82}  \\
				$nclt\_6$  & 12.21         & $\times$              & \textbf{2.17}  \\
				$nclt\_7$  & 1.42          & \textbf{1.21}  & 1.38           \\
				$nclt\_8$  & 2.57          & $\times$              & \textbf{2.20}  \\
				$nclt\_9$  & 3.54          & $\times$              & \textbf{2.22}  \\
				$nclt\_10$ & 2.56          & $\times$              & \textbf{2.24}  \\
				$nclt\_11$ & \textbf{2.14} & $\times$              & 2.25           \\
				$nclt\_12$ & 2.37          & $\times$              & \textbf{2.23}  \\
				$nclt\_13$ & 4.39          & $\times$              & \textbf{2.13}  \\
				$nclt\_14$ & 2.24          & $\times$              & \textbf{2.19}  \\
				$nclt\_15$ & 5.44          & $\times$              & \textbf{2.16}  \\
				$nclt\_16$ & 0.95          & \textbf{0.92}  & \textbf{0.92}  \\
				$utbm\_1$  & 16.89         & \textbf{8.78}  & 15.27          \\
				$utbm\_2$  & 16.49         & \textbf{12.81} & 18.58          \\
				$utbm\_3$  & 10.65         & $\times$              & \textbf{10.16} \\
				$utbm\_4$  & 13.88         & $\times$              & \textbf{13.44} \\
				$utbm\_5$  & 13.20         & $\times$              & \textbf{11.71} \\
				$utbm\_6$  & 13.53         & $\times$              & \textbf{13.40} \\
				$utbm\_7$  & 13.02         & $\times$              & \textbf{12.15} \\
				$ulhk\_1$  & 1.54          & 1.03           & \textbf{0.93}  \\
				$ulhk\_2$  & 3.37          & \textbf{3.15}  & 3.24           \\
				$kaist\_1$ & 1.55          & $\times$              & \textbf{0.86}  \\
				$kaist\_2$ & 3.61          & $\times$              & \textbf{3.44}  \\
				$kaist\_3$ & 76.94         & $\times$              & \textbf{1.04}  \\ \hline
			\end{tabular}
		\end{threeparttable}
		\begin{tablenotes}
			\footnotesize
			\item[] \textbf{Denotations}: “$\times$” means the system fails to run entirety on the corresponding sequence.
		\end{tablenotes}
	\end{center}
\end{table}

We examine the effectiveness of our proposed semi-elastic state estimation on accuracy by comparing the ATE result utilizing the traditional optimization-based LiDAR-inertial state estimation, the elastic optimization-based LiDAR-inertial state estimation and the semi-elastic optimization-based LiDAR-inertial state estimation. Results in Table \ref{table4} demonstrate that the proposed semi-elastic optimization-based LiDAR-inertial state estimation can achieve more accurate and robust pose estimation than the traditional and the elastic optimization-based LiDAR-inertial state estimation on most sequences. Especially, the elastic method cannot run entirely on most sequences, demonstrating the extremely poor robustness of this scheme.

\begin{figure}
	\begin{center}
		\includegraphics[scale=0.36]{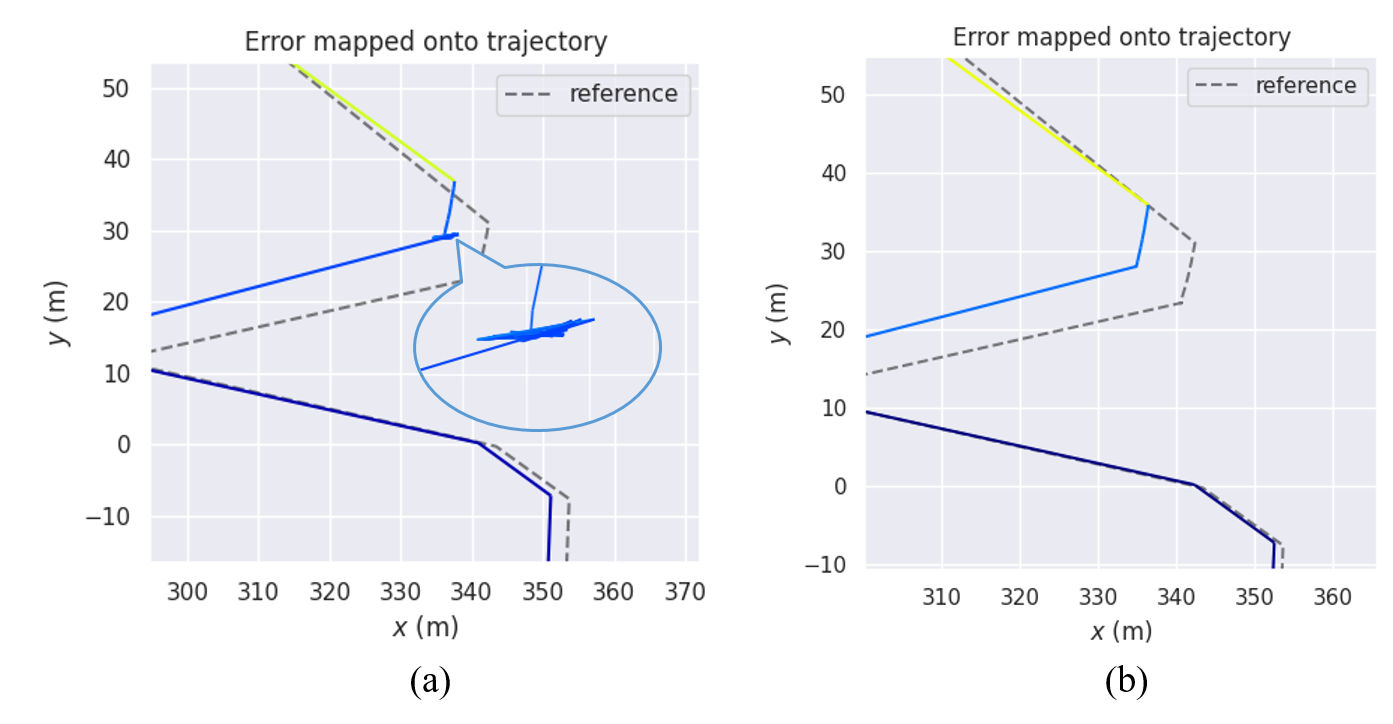}
		\caption{(a) The local zigzag of trajectory estimated by traditional LiDAR-inertial state estimation. (b) The smooth trajectory estimated by our semi-elastic LiDAR-inertial state estimation.}
		\label{fig3}
	\end{center}
\end{figure}

In addition, we utilize $utbm\_6$ as the examplar sequence to examine the effectiveness of our method on consistency by comparing the smoothness of estimated trajectory and the curve of estimated velocity. In theory, the estimated trajectory should be smooth but not zigzag. If there is a zigzag somewhere, the consistency of estimated state there is very poor. As illustrated in Fig. \ref{fig3}, the local trajectory with zigzag shape is easy to appear utilizing traditional optimization-based LiDAR-inertial state estimation method, which means the strong local inconsistency. By contrast, our semi-elastic optimization-based LiDAR-inertial state estimation can obtain a smooth trajectory. The trajectory corresponding to elastic optimization-based LiDAR-inertial state estimation drifts directly, so we omit its visualization in Fig. \ref{fig3}.

\begin{figure}
	\begin{center}
		\includegraphics[scale=0.47]{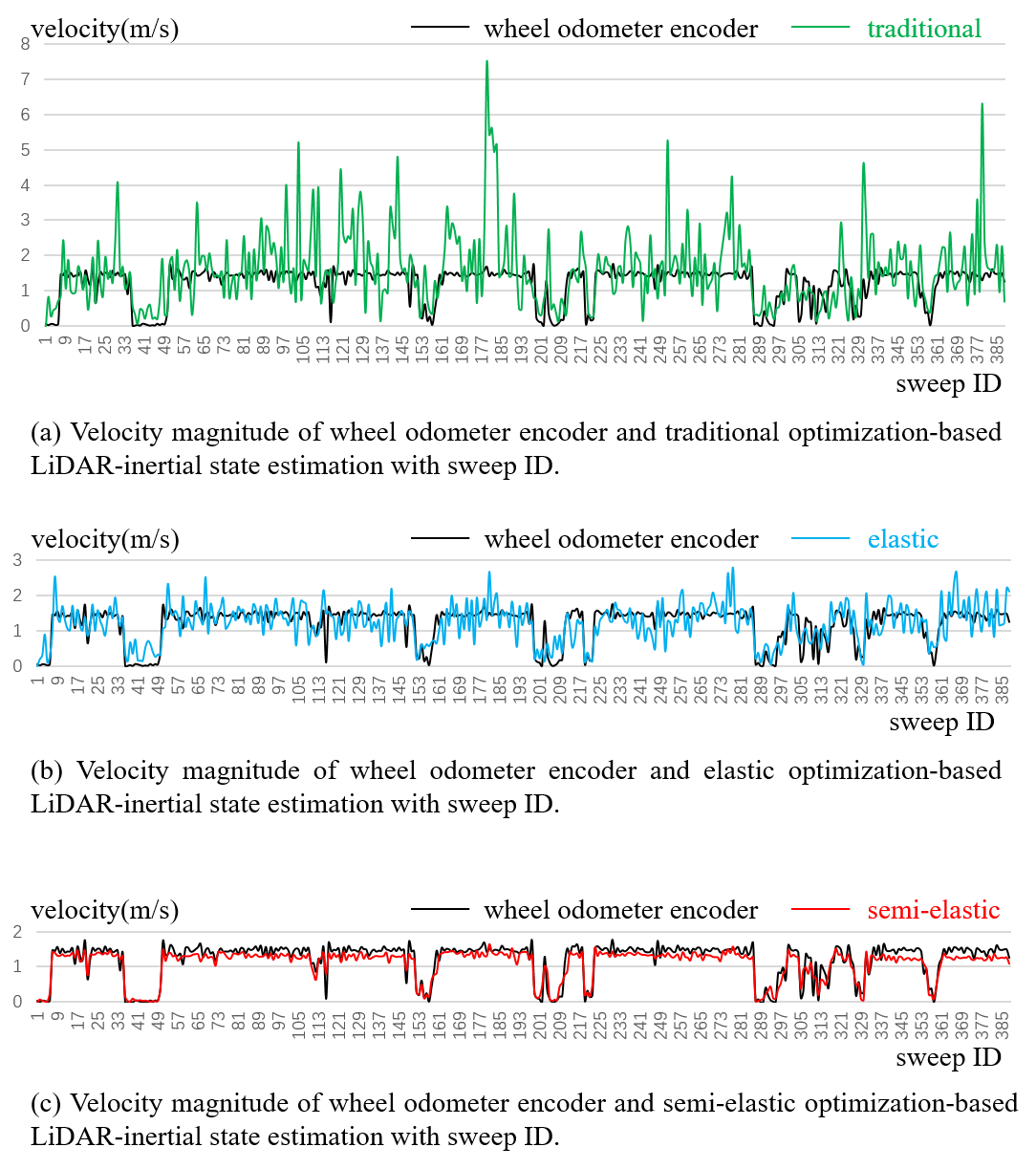}
		\caption{The curve of velocity magnitude estimated by three state estimation methods and measured by wheel odometer encoder on $nclt\_16$. We downsampled the sample size by 1/20 when plotting.}
		\label{fig4}
	\end{center}
\end{figure}

In practice, it is difficult for us to obtain the ground truth value of velocity. However, we can still make a qualitative assessment for the accuracy of velocity. In theory, the velocity of a moving vehicle should be continuous and smooth, but not oscillating at a high frequency. Therefore, the smoother the curve of a velocity function with respect to time, the more accurate the velocity is. We utilize the sequence $nclt\_16$ as example. As illustrated in Fig. \ref{fig4}, compared to the high frequency oscillation curve of traditional and elastic optimization-based LiDAR-inertial state estimation, the curve of proposed semi-elastic optimization-based LiDAR-inertial state estimation is much smoother. That means there is still a gap in consistency even if the ATE of estimated pose by the three methods is similar (recorded in Table \ref{table4}). In addition, although the groundtruth of velocity vector cannot be obtained, the $nclt$ dataset has the observation from wheel odometer encoder to provide the magnitude of velocity for reference, which is also in good agreement with the velocity curve estimated by our method. This shows that our method can not only ensure the consistency, but also greatly improve the accuracy of estimated velocity, demonstrating the effectiveness of semi-elastic optimization-based LiDAR-inertial state estimation method.

\subsection{Time Consumption}
\label{Time Consumption}

\begin{table}[]
	\begin{center}
		\caption{Time Consumption Per Sweep (unit: ms)}
		\label{table5}
		\begin{tabular}{c|cc|c}
			\hline
			& \begin{tabular}[c]{@{}c@{}}Semi-Elastic\\ State Estimation\end{tabular}   & Point Registration & Total \\ \hline
			$nclt\_1$  & 47.06                         & 11.56              & 61.10 \\
			$nclt\_2$  & 59.19                         & 9.24               & 70.88 \\
			$nclt\_3$  & 55.55                         & 10.26              & 67.55 \\
			$nclt\_4$  & 59.74                         & 5.82               & 68.23 \\
			$nclt\_5$  & 46.12                         & 12.75              & 61.10 \\
			$nclt\_6$  & 46.67                         & 10.71              & 59.08 \\
			$nclt\_7$  & 52.60                         & 6.20               & 61.22 \\
			$nclt\_8$  & 52.44                         & 12.03              & 66.40 \\
			$nclt\_9$  & 54.07                         & 11.02              & 66.97 \\
			$nclt\_10$ & 52.20                         & 9.13               & 62.10 \\
			$nclt\_11$ & 48.18                         & 11.68              & 61.45 \\
			$nclt\_12$ & 48.41                         & 11.57              & 61.69 \\
			$nclt\_13$ & 47.70                         & 11.41              & 60.76 \\
			$nclt\_14$ & 46.69                         & 9.67               & 57.86 \\
			$nclt\_15$ & 50.79                         & 10.14              & 65.27 \\
			$nclt\_16$ & 52.11                         & 2.63               & 57.06 \\
			$utbm\_1$  & 47.55                         & 11.65              & 60.71 \\
			$utbm\_2$  & 46.90                         & 11.82              & 60.20 \\
			$utbm\_3$  & 47.27                         & 9.57               & 58.47 \\
			$utbm\_4$  & 48.73                         & 11.61              & 61.78 \\
			$utbm\_5$  & 48.57                         & 11.07              & 61.06 \\
			$utbm\_6$  & 48.59                         & 11.23              & 61.27 \\
			$utbm\_7$  & 46.02                         & 12.01              & 59.27 \\
			$ulhk\_1$  & 40.09                         & 1.43               & 43.09 \\
			$ulhk\_2$  & 26.05                         & 1.54               & 29.26 \\
			$kaist\_1$ & 31.14                         & 6.73               & 38.73 \\
			$kaist\_2$ & 28.16                         & 4.99               & 34.00 \\
			$kaist\_3$ & 55.42                         & 6.98               & 63.36 \\ \hline
		\end{tabular}
	\end{center}
\end{table}

We evaluate the runtime breakdown (unit: ms) of our system for all sequences. In general, the most time-consuming modules are the semi-elastic state estimation module, and the point registration module. Therefore, for each sequence, we test the time cost of above two modules, and the total time for handling a sweep. Results in Table \ref{table5} show that our system takes 60$\sim$70\,ms to handle a sweep, while the time interval of two consecutive input sweeps is 100\,ms. That means our system can not only run in real time, but also save 30$\sim$40\,ms per sweep.

\subsection{Visualization for map}
\label{Visualization for map}

\begin{figure}
	\begin{center}
		\includegraphics[scale=0.24]{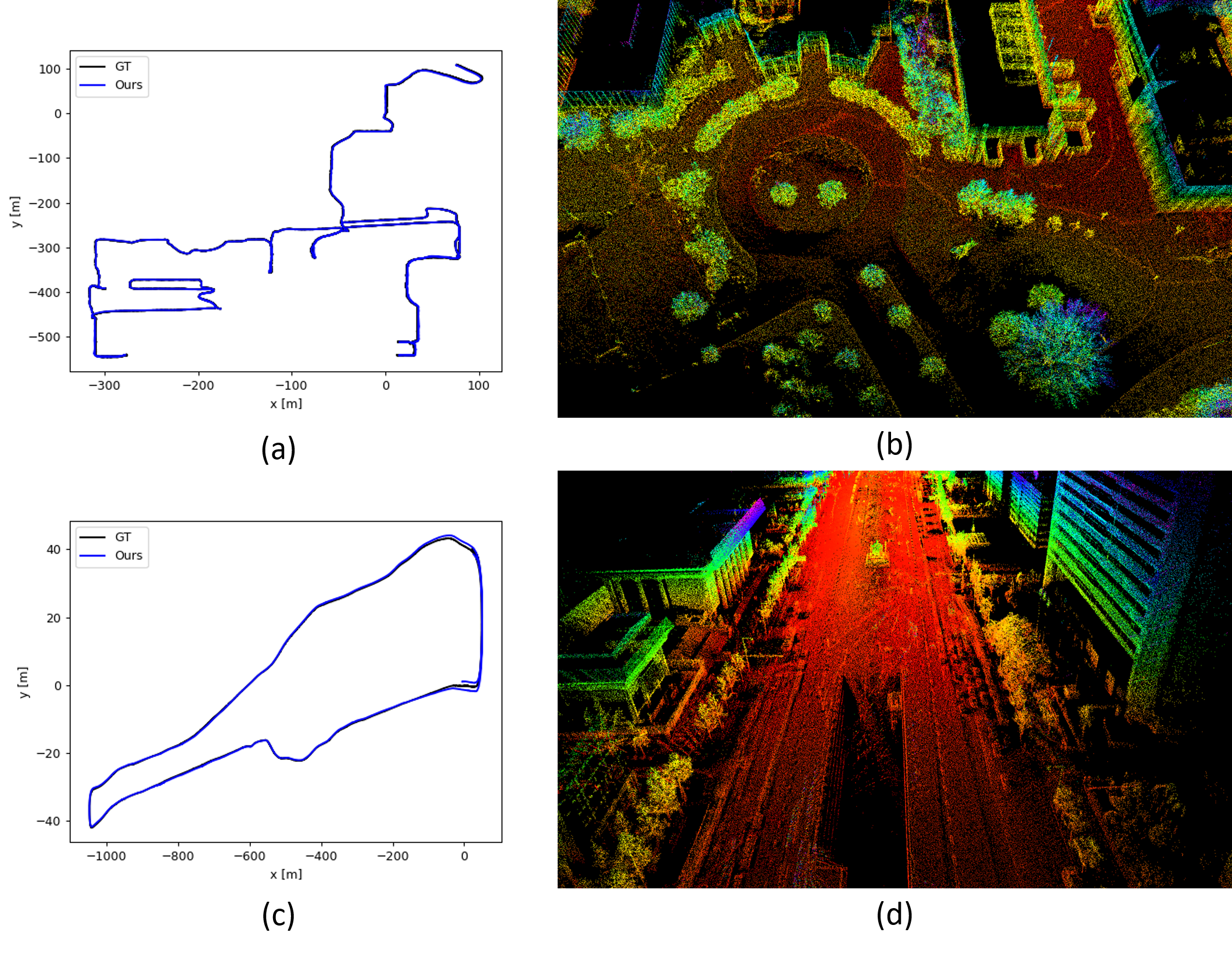}
		\caption{(a) and (c) are the comparison results between our estimated trajectories and ground truth on the exemplar sequences $nclt\_7$ and $kaist\_3$. (b) and (d) are the local point cloud map of $nclt\_7$ and $kaist\_3$ respectively.}
		\label{fig5}
	\end{center}
\end{figure}

We visualize the trajectories and local point cloud maps estimated by our system. The comparison result between our estimated trajectory and ground truth of the exemplar sequences $nclt\_7$ and $kaist\_3$ is shown in Fig. \ref{fig5} (a) and (c), where our estimated trajectories and ground truth almost exactly coincide. Fig. \ref{fig5} (b) and (d) show sufficient accuracy for some local structures, where the distribution of the points is also uniform. The demo video of our supplementary material demonstrates that our system can also estimate accurate pose using 16-line Robosense LiDAR.

\section{Conclusion}
\label{Conclusion}

This paper proposes the semi-elastic optimization-based LiDAR-inertial state estimation method, which can give the state enough flexibility to be optimized to the correct value, thus preferably ensure the accuracy, consistency and robustness of state estimation. We embed the proposed LiDAR-inertial state estimation method into an optimization-based LIO framework.

Our system achieves state-of-the-art accuracy and robustness on four public datasets. Meanwhile, the ablation study demonstrates that the proposed semi-elastic LiDAR-inertial method can achieve better consistency for the estimated state.

\bibliographystyle{IEEEtrans}
\bibliography{IEEEabrv,IEEEExample}




\ifCLASSOPTIONcaptionsoff
  \newpage
\fi

\end{document}